\pdfoutput=1
\documentclass[11pt,a4paper]{article}
\usepackage{authblk}
\usepackage[hyperref]{emnlp2020}
\usepackage{times}
\usepackage{latexsym}

\usepackage{todonotes}
\usepackage{enumitem}
\usepackage{amssymb}
\usepackage{tabularx}
\usepackage{subcaption}
\usepackage{booktabs}
\usepackage{multirow}
\usepackage{makecell}

\usepackage{microtype}

\aclfinalcopy 
\def\aclpaperid{2281} 

\title{Spot The Bot: A Robust and Efficient Framework for the Evaluation of Conversational Dialogue Systems}

\author[1]{Jan Deriu}
\author[1]{Don Tuggener}
\author[1]{Pius von D{\"a}niken}
\author[3]{Jon Ander Campos}
\author[2]{\\ Alvaro Rodrigo} 
\author[4]{Thiziri Belkacem}
\author[3]{Aitor Soroa}
\author[3]{Eneko Agirre}
\author[1]{Mark Cieliebak}

\affil[1]{Zurich University of Applied Sciences (ZHAW), Winterthur, Switzerland, \textit{\{deri, tuge, vode, ciel\}@zhaw.ch}}
\affil[2]{National Distance Education University (UNED), Madrid, Spain, \textit{alvarory@lsi.uned.es}}
\affil[3]{University of the Basque Country (UPV/EHU), Donostia, Spain, \textit{\{jonander.campos, a.soroa, e.agirre\}@ehu.eus}}
\affil[4]{Synapse Développement, Toulouse, France, \textit{belkacemthiziri@gmail.com}}

\date{}

\begin{document}
\maketitle
\begin{abstract}
The lack of time-efficient and reliable evaluation methods hamper the development of conversational dialogue systems (chatbots). Evaluations requiring humans to converse with chatbots are time and cost-intensive, put high cognitive demands on the human judges, and yield low-quality results.
In this work, we introduce \emph{Spot The Bot}, a cost-efficient and robust evaluation framework that replaces human-bot conversations with conversations between bots. 
Human judges then only annotate for each entity in a conversation whether they think it is human or not (assuming there are humans participants in these conversations). 
These annotations then allow us to rank chatbots regarding their ability to mimic the conversational behavior of humans. 
Since we expect that all bots are eventually recognized as such, we incorporate a metric that measures which chatbot can uphold human-like behavior the longest, i.e., \emph{Survival Analysis}. 
This metric has the ability to correlate a bot's performance to certain of its characteristics (e.g., \ fluency or sensibleness), yielding interpretable results.
The comparably low cost of our framework allows for frequent evaluations of chatbots during their evaluation cycle. We empirically validate our claims by applying \emph{Spot The Bot} to three domains, evaluating several state-of-the-art chatbots, and drawing comparisons to related work. The framework is released as a ready-to-use tool.
\end{abstract}

\section{Introduction}\label{sec:intro}
Evaluation is a long-standing issue in developing conversational dialogue systems (i.e., \ chatbots). The underlying difficulty in evaluation lies in the problem's open-ended nature, as chatbots do not solve a clearly-defined task whose success can be measured in relation to an a priori defined ground truth. Automatic metrics have so far failed to show high correlation with human evaluations ~\cite{liu-etal-2016-evaluate,lowe-etal-2017-towards,mehri2020usr}.
Human evaluation approaches are mainly classified according to the following: single-turn vs. multi-turn evaluation, and direct user evaluation vs. expert judge evaluation. Single-turn analysis is usually performed by a human judge that rates a single response of the bot to a given context, whereas multi-turn analysis is often performed by a user that interacts with the bot and rates the interaction. 
Single-turn ratings disregard the multi-turn nature of a dialogue \cite{see-etal-2019-makes}. Although more and more multi-turn evaluations are performed, most of them are based on human-bot conversations, which are costly to obtain and tend to suffer from low quality \cite{dinan2020convai2}.
The instructions to be followed by annotators are often chosen ad-hoc and there are no unified definitions. Compounded with the use of often criticized Likert scales \cite{amidei2019agreement}, these evaluations often yield a low agreement. The required cost and time efforts also inhibit the widespread use of such evaluations, which raises questions on the replicability, robustness, and thus significance of the results. \\
In this work, we present the \emph{Spot The Bot} framework, a cost-efficient evaluation methodology that can be used to rank several bots with regard to their ability to disguise as humans. It works as a multi-turn-based evaluation with human judges. \emph{Spot The Bot} is based on two observations: First, chatbots are trained on conversations between humans, and thus, they should be evaluated regarding their ability to mimic human behavior. Second, the longer a conversation is, the more likely it is that a bot exhibits non-human-like behavior. \\
\emph{Spot The Bot} works by  generating conversations between bots, then mixing these bot-bot conversations with human-human conversations and letting human judges decide for each entity in the conversations if it is a human or a bot. The conversations are rated at different points in time, which introduces the time-dependent component. This setting allows for two different analyses: a \emph{ranking based on pairwise comparisons of bots}, and the application of the \emph{Survival Analysis}, which computes the survival rate for each bot at different conversation lengths. 
Furthermore, the human judges annotate the entities with respect to more fine-grained features, which can be chosen based on characteristics that the bots are expected to exhibit (e.g. fluency or informativeness). The Survival Analysis further allows to pin down the features that contribute to a dialogue system's survival, enabling interpretable results. \\
We show that our framework produces reliable, repeatable results, while being quicker and more cost-effective to run than related approaches, as it does not rely on human-bot conversations and generally requires fewer annotations. Furthermore, we show that disagreement between human annotators can be interpreted as a feature of a system's performance, rather than a weakness in the evaluation approach. We apply the framework to three well-known domains and common baselines and state-of-the-art systems to produce a stable ranking among them. We release the framework as a ready-to-use tool for evaluating dialogue systems into which different systems can be plugged and compared\footnote{\url{https://github.com/jderiu/spot-the-bot-code}}.

\section{Related Work}
\label{sota}
There exist various methods to evaluate dialogue systems, both automatic and human-based, but no single evaluation metric is widely agreed upon in the scientific community \cite{deriu2019survey}. 
Automatic evaluation metrics for chatbots are known to correlate poorly with human ratings~\cite{liu-etal-2016-evaluate,lowe-etal-2017-towards,mehri2020usr}, so we focus on human-based approaches, which can be classified in two dimensions: 1) single-turn vs. multi-turn approaches, and 2) approaches where the dialogue systems are judged by the user directly (interactive) or where judgments are made by objective experts, who do not participate in the dialogue (static). \\
\noindent{\bf Single-turn Static Evaluations.} Evaluations based on a static context and a single response from the dialogue systems are widely adopted. Usually, the rating is performed by expert raters that read the response of one or more dialogue systems to a static context and rate the responses~\cite{Galley2018EndtoEndCM}. Alternatively, the responses of two bots can be compared directly to choose a preferred answer \cite{li-etal-2016-persona}. While being relatively time and cost-efficient, single-turn evaluation fails to capture the conversation's quality as a whole. A system that tends to produce repeated answers can obtain a high single-turn score, albeit a low multi-turn one \cite{see-etal-2019-makes}. Some authors also report poor inter-annotator agreement \cite{ghandeharioun2019approximating}.
\begin{figure*}[ht!]
    \centering
    \includegraphics[width=0.95\textwidth]{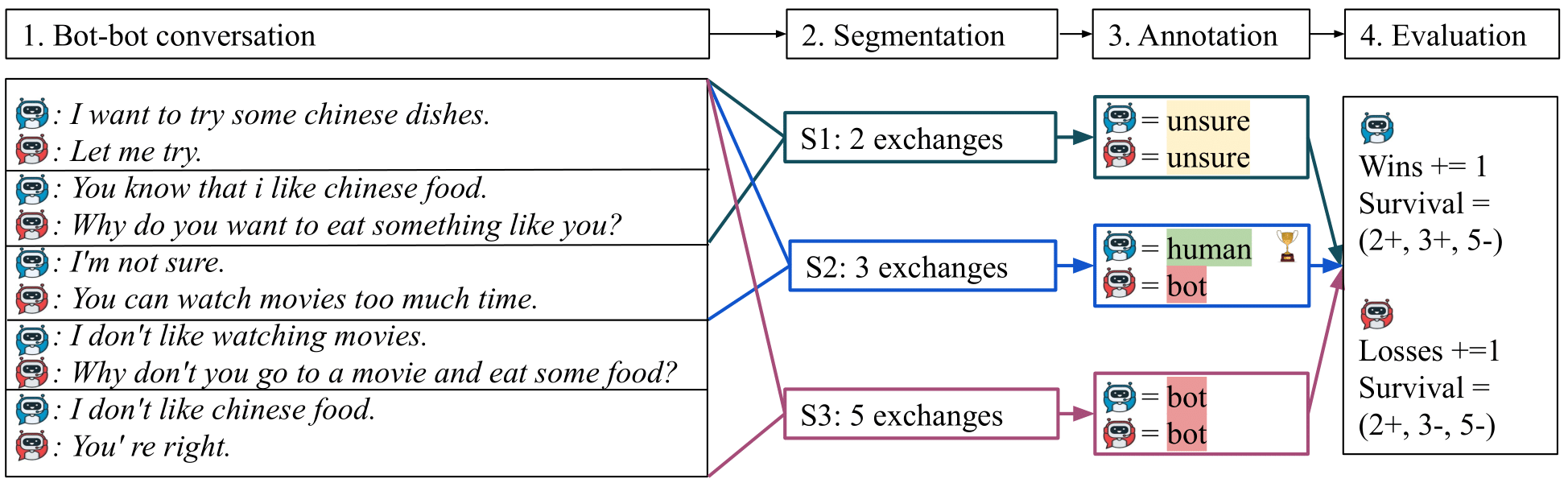}
    \caption{Overview of the \emph{Spot The Bot} process for one conversation. 1: A bot-bot conversation is segmented into different lengths (e.g. 2, 3, and 5 exchanges). 2: These segments are shown to distinct sets of annotators who judge whether each entity is a bot. 4: The winner is determined for each annotated segment and survival statistics are updated. This process is repeated for all conversations between the competing bots.}
    \label{fig:example}
\end{figure*}
\\

\noindent{\bf Human-Bot Conversations.}  In order to perform interactive multi-turn evaluations, the standard method is to let humans converse with a chatbot and rate it afterward \citep{ghandeharioun2019approximating}, typically using Likert scales ~\cite{van-der-lee-etal-2019-best}. The ConvAI2 challenge~\cite{convai12} and the Alexa Prize~\cite{venkatesh2018evaluating} applied this procedure. Apart from the high cost of collecting human-bot conversations, this approach puts a high cognitive strain on humans, as they have to perform several tasks at once~\cite{SCHMITT201512}. Besides, it is not always possible to get sensible conversations with bots, making it hard to get high-quality conversations. In fact, in the ConvAI2 challenge, half of the collected human-bot conversations were discarded due to their low quality~\cite{convai12}. Finally, Likert scales are known to suffer from high annotation variance~\cite{ghandeharioun2019approximating}, require normalization a posteriori, are prone to order effects and are less reliable than ranking-based ratings~\cite{amidei-etal-2019-use}. 
\\

\noindent{\bf Self-talk.} Recently, using self-talk dialogues, i.e., \ dialogues where a bot talks to itself, gained traction as a cost-effective basis for evaluation. This idea is closely related to user simulations used to evaluate task-oriented systems \cite{schatzmann:2006:us_survey}. \newcite{ghandeharioun2019approximating} and \newcite{deriu-cieliebak-2019-towards} use self-talk to produce automatic evaluations. In ACUTE-EVAL \cite{li2019acuteeval}, the authors propose to let humans evaluate self-talk dialogues. Since self-talk does not allow direct comparisons between bots, the authors let humans read two self-talk conversations side-by-side and rate them with respect to various features.
This increases the cognitive complexity of the annotation task. 
Furthermore, the resulting ranking of the bots is per criterion, whereas our method produces one ranking and can optionally incorporate annotations of features that yield interpretability of the results. \\

\noindent{\bf Turing Test.} \emph{Spot The Bot} is reminiscent of the Turing Test \cite{TuringTest}, as the dialogue systems are evaluated based on their ability to mimic human behavior. The Turing test served as a useful mental model for understanding what machine intelligence might mean. However, it has also been criticized as a way to identify intelligence in NLP systems. \citet{bender-2020-acl} argues that a system may fool a user into believing it is human, and yet this does not prove that the system understands the meaning of the conversation they are having. In our approach, we claim that \emph{failing} the test is a valid indicator to discriminate among bots. In fact, we presume that eventually all bots will fail the test, and we collect a time component to record the time it takes for a bot to be detected.

\section{Spot The Bot}

In this section, we first provide an overview of the \emph{Spot The Bot} framework and then describe the evaluation process's individual steps.

\subsection{Overview}
\label{sec:setting}

\emph{Spot The Bot} employs a tournament among chatbots to determine which performs the best at mimicking humans' conversational behavior. To measure the success of each bot, human crowdworkers are shown conversations between two competing bots at a time, mixed with conversations between two humans. The crowdworkers' task is to determine for each entity in a conversation whether it is a human or a bot (or whether the crowdworker is unsure). The bot that is most frequently annotated as being human wins the tournament. Figure \ref{fig:example} provides an overview of the process for one conversation. \\
There are different use cases for \emph{Spot The Bot}, e.g., \ when a novel dialogue strategy is to be compared against existing ones or if a set of chatbots is to be ranked in the context of a shared task. On top of returning a ranking, \emph{Spot The Bot} employs the Survival Analysis, which introduces a time aspect into the evaluation and provides insights into how different features correlate to the bots' ability to pass as a human. \\
Formally, assume a pool of $b$ bots $\{B_1, ..., B_b\}$, which is to be ranked. For each pair of bots, a set of conversations is sampled by letting the bots talk to each other, where $S_{ij}$ denotes the set of conversations between bots $B_i$ and $B_j$. Each conversation is defined as a sequence of exchanges $e_0, ..., e_N$, where each exchange consists of two turns: $e_i = \{t_0^{e_i}, t_1^{e_i}\}$, one for each entity. \\

\noindent{\bf Segmentation.} The more exchanges there are in a conversation, the more likely it is that a bot gets recognized as such. Thus, we show different segments of the conversation to the crowdworkers. A segment is defined as the first $k$ exchanges of the dialogue: $S_{ij}^k = e_0, ..., e_k$. Thus, an annotator only sees the first $k$ exchanges of the conversation.\footnote{We experimented with letting crowdworkers decide where they were sure that an entity is a bot or a human. However, this approach required too much fine-tuning to constrain erratic annotator behavior, cf.\ Appendix \ref{sec:appendix_gamification}.}  Each segment of the same conversation is rated by a different annotator to avoid that one annotator sees parts of the same conversation multiple times, which would bias the rating. We choose different segment lengths since we cannot know a priori which length is sufficient for the different bots to be recognized as such.\\

\noindent{\bf Human Conversations.} We add conversations among humans to the pool of conversations that are to be rated. The human conversations are sampled from the training set used to train the dialogue systems in the respective domain. The results of the annotations of the human dialogues establish an upper bound for the evaluation. Also, they are meant to prevent annotators from concluding that all entities are bots.\footnote{We investigated if annotators realize that conversations are either between bots or humans by looking at ratios of conversations where both entities are labeled identically, but found no evidence that this happens more often than by chance.} \\

\noindent{\bf Annotation.} The annotation procedure works in two steps: First, the annotators have to decide for each entity in a conversation segment if it is a bot or a human. Second, to correlate the outcome to various characteristics of a bot, the framework allows rating specific features (e.g., \ fluency or appropriateness). The framework then measures the influence of these features on the survival time of the bots, which serves as an explainability component (cf. Sections \ref{sec:survival} and \ref{sec:survival_results}).

\noindent{\bf Features.} We chose three features: sensibleness, specificity \cite{adiwardana2020towards}, and fluency. The first two are shown to capture the core conversational behavior of answering sensibly and not with illogical statements while being specific to the conversation's given context. The third feature states if the utterances are grammatically correct and fluent. The features are rated by preference ranking, that is, the annotator states which of the two entities performed better with respect to the features.  \\

\subsection{Ranking}
We define a win function for the annotations of the pairwise, direct conversations between two bots. The outputs of the win function are aggregated to determine the overall winner of the tournament. \\

\noindent{\bf Win Function.} Each annotation at each segment length $S_{ij}^k = e_0, ... , e_{k}$ of a conversation constitutes the result of one annotation applied by one crowdworker, individually labeling each of the two entities as either \textit{bot, human}, or \textit{unsure}. The winner of segment $S_{ij}^k$ under a crowdworker's annotation is determined by the following ordering of the labels: $\textit{human} >  \textit{unsure} > \textit{bot}$. That is, if bot $B_i$ is assigned the label \textit{human} and bot $B_j$ has label \textit{bot} or \textit{unsure}, $B_i$ has won the segment.\footnote{This process is repeated for all crowdworkers who annotated the segment - in our case two per segment - and each win is counted separately.} Similar to \newcite{bojar-etal-2013-findings},  we define a win rate of $B_i$ against $B_j$ to aggregate the wins from all segments of all annotations stemming from conversations between bots $B_i$ and $B_j$, as: 

\begin{equation}\label{eq:win_function}
    \frac{\textsc{wins($B_i, B_j$)}}{\textsc{wins($B_i, B_j$)} + \textsc{wins($B_j, B_i$)}}
\end{equation}
where $\textsc{wins($B_i$, $B_j$)}$ denotes the number of times that $B_i$ wins against $B_j$. \\

\noindent{\bf Ranking.} To create the ranking, we follow the approach by \newcite{dusek-etal-2018-findings}, where the ranking is generated by the TrueSkill \cite{herbrich2006trueskill} algorithm based on the win rate, and significant differences in performance are determined by bootstrap sampling. The result is a ranked set of clusters, where each cluster is composed of entities that do not have a significant difference in performance. 

\subsection{Survival Analysis}\label{sec:survival}
While pair-wise win rates are well-suited to provide a \textit{relative} ranking among a pool of bots, it does not serve as an \textit{absolute} evaluation of a single bot's ability to disguise as a human. Also, the conversations' segmentation introduces a time component, which we leverage to investigate our intuition that bots are more likely to reveal themselves in longer conversations. In our evaluation, a bot that is able to disguise in long conversations can be said to be most successful.
Thus, we complement our evaluation with \emph{Survival Analysis}. \\
Survival Analysis estimates probabilities for the occurrence of an event at different points in time. It has a long history in the medical domain, where it is used to estimate the effectiveness of different treatments \cite{li2013survival}. In engineering disciplines, it is applied to estimate the time to failure of machine components \cite{Eyal2014}. In our case, we are interested in the time, corresponding to the number of exchanges, until a dialogue system is spotted as such. 
In addition, Survival Analysis allows us to correlate finer-grained characteristics to the survival probability, which allows us to inspect which of the annotated features impact a bot's survival. \\
We interpret the annotation data as such: the \emph{spotted} event occurred if the system was annotated as ``bot'' and it \emph{survived} if it was annotated as ``unsure'' or ``human''. Let $k$ be the number of exchanges in the annotated conversation segment, meaning that each dialog system produced $k$ outputs. If the dialog system was not spotted, we know it survived for at least $k$ exchanges. This is a so-called right-censored data point. If the dialogue system was spotted as such, we cannot tell the exact number of exchanges it took for an annotator to spot it, meaning it could have taken less than $k$ exchanges. We thus record that the spotting event happened in the interval $(0, k]$, a so-called interval-censored event. \\
From this data, we can get non-parametric estimates of the survival function of the different systems per domain \citep{nonparametric_survival_censored}.  To check whether these differences are significant, we apply a generalized log-rank test \citep{generalized_logranktest_zhao2004}. We use the \emph{Cox Proportional Hazards Model} \citep{cox_ph_model} to study the influence of the features outlined in Section \ref{sec:setting} on the time before the systems are spotted.\footnote{We use the \emph{icenReg} R package \citep{icenReg}, which allows us to fit a Cox model to our interval-censored data.}

\section{Experiments}
{\bf Domains.} We apply \emph{Spot The Bot} to three widely used domains for conversational dialogue systems:  Dailydialog \cite{li-etal-2017-dailydialog}, Empathetic Dialogues \cite{rashkin-etal-2019-towards}, and PersonaChat \cite{zhang-etal-2018-personalizing}. For each domain\footnote{See details in Appendix \ref{app:domain}.}, we prepared a pool of bots to be ranked and analyzed. For each pair of bots, we sampled $|S_{ij}| = 45$ conversations. For this, we seed the conversations by using the first exchange of a conversation in the test set, which is sampled at random. Although there exists a probability that the bots resample parts of a conversation, we did not find evidence of this happening. In fact, only 2\% of all sampled conversations contain an exchange, which can be found in the training material. For the annotation task, we recruited paid crowdworkers from Amazon Mechanical Turk (AMT). To avoid that, the results are biased towards the performance of a few crowdworkers, we designed a Human Intelligence Task as a batch of 20 conversations, and each worker was only allowed to work on three batches. We designed the batches so that two segments of the same conversations never appear in the same batch, and each batch contains different segments of different conversations.  \\

\noindent{\bf Segmentation.} The segment lengths are based on the lengths of the dialogues in a domain. Since we add human conversations of the training set to be rated, the sampled dialogues should adhere to their lengths. PersonaChat and Dailydialog have longer conversations; thus, we used segments of 2, 3, and 5 exchanges. The Empathetic Dialogue domain has shorter dialogues; thus, we used segment lengths of 1, 2, and 3 exchanges.\\

\noindent{\bf Dialogue Systems.} For each domain, we prepared a pool of dialogue systems to be ranked. If applicable, we reused existing systems. 
In order to assess the performance of \emph{Spot The Bot} regarding weak models, we trained a small sequence-to-sequence model (DR) for only 3 epochs, which returns mostly general answers. 
For the Dailydialog domain, we trained all bots in the pool using ParlAI as there were no pre-trained models available. To leverage the recently developed language models, we fine-tune a GPT-2 (GPT) model~\cite{radford2018gpt}, and a BERT-Rank (BR) model. Additionally, we train a sequence-to-sequence model (S2) with attention to compare the language models to previous state-of-the-art approaches. 
Together with the DR model, the pool consists of $b=4$ systems.
For the Empathetic Dialogues, we prepared the same pool of models as in Dailydialog. Since the recently developed Blender model~\cite{roller2020recipes} is trained on the Empathetic Dialogue dataset as well, we add the pre-trained version to the pool (BL). 
For the PersonaChat domain, we mostly reuse the openly available systems of the ConvAI2 challenge~\cite{dinan2020convai2}, namely, Lost in Conversation\footnote{\url{https://github.com/atselousov/transformer\_chatbot}} (LC) and Huggingface~\footnote{\url{https://github.com/huggingface/transfer-learning-conv-ai}} (HF), which were the top-rated dialogue systems in the ConvAI2 challenge \cite{dinan2020convai2}, as well as KVMemNN (KV), which served as the baseline. We also add the Blender model, which is also trained in this domain. In order to have more retrieval based systems, we train a BertRank (BR) model. Together with the DR model, the pool consists of $b = 6$ different dialogue systems.

\subsection{Ranking Results}
Table \ref{tab:win_rates} gives an overview of the win rates for each pair of bots and their ranking ranges. The Chi-square test computes the significance. For each domain, most pairwise win-rates are significant. \\
\begin{table}[h!]
\centering
\scalebox{0.70}{
\begin{tabular}{l|cccccc|cc} 
\multicolumn{9}{c}{Dailydialog} \\
\toprule
\textsc{} & \textsc{GPT} &  \textsc{BR} &  \textsc{S2} & \textsc{DR} & & & \textsc{WR}& \textsc{Range}\\
\hline 
\textsc{GPT} & -             & \textbf{0.67} & \textbf{0.77} & \textbf{0.93} & & & 0.79 & (1,1) \\
\textsc{BR} & \textbf{0.33} & -          & \textbf{0.79}    & \textbf{0.83} & & & 0.65 & (1,2)  \\
\textsc{S2} & \textbf{0.23} & \textbf{0.21} & -             & \textbf{0.74} & & & 0.39 & (3,3) \\
\textsc{DR} & \textbf{0.07} & \textbf{0.17} & \textbf{0.26} & -             & & & 0.16 & (4,4) \\
\bottomrule
\multicolumn{9}{c}{Empathetic Dialogues} \\
\toprule
\textsc{} & \textsc{BL}& \textsc{BR} &  \textsc{GPT} &  \textsc{S2} & \textsc{DR} & & \textsc{WR} & \textsc{Range}\\
\hline 
\textsc{BL}& - & \textbf{0.82} & \textbf{0.83} & \textbf{0.9} & \textbf{0.94} && 0.87 & (1,1)  \\
\textsc{BR}& \textbf{0.18} & - & 0.51 & \textbf{0.77} & \textbf{0.93} && 0.59 & (2,3)  \\
\textsc{GPT}& \textbf{0.17} & 0.49 & - & \textbf{0.61} & \textbf{0.73} && 0.50 & (2,3) \\
\textsc{S2}& \textbf{0.10} & \textbf{0.23} & \textbf{0.39} & - & \textbf{0.63} && 0.33 & (4,4) \\
\textsc{DR}& \textbf{0.06} & \textbf{0.07} & \textbf{0.27} & \textbf{0.37} & - && 0.19  & (5,5)\\
\bottomrule

\multicolumn{9}{c}{PersonaChat} \\
\toprule
\textsc{} & \textsc{BL}& \textsc{LC} &  \textsc{KV} &  \textsc{HF} & \textsc{BR}  & \textsc{DR} &\textsc{WR} & \textsc{Range}\\
\hline 
\textsc{BL}& - & 0.56 & \textbf{0.68} & \textbf{0.72} & \textbf{0.84} & \textbf{0.95}&  0.75 & (1-1) \\
\textsc{LC}& 0.44 & - & 0.54 & \textbf{0.72} & \textbf{0.75} & \textbf{0.89} & 0.69  & (2-3)\\
\textsc{KV}& \textbf{0.32} & 0.46 & - & \textbf{0.77} & \textbf{0.74} & \textbf{0.91} & 0.64 & (2-3)\\
\textsc{HF}& \textbf{0.28} & \textbf{0.28} & \textbf{0.23} & - & \textbf{0.63}& \textbf{0.89} & 0.46& (4-4)\\
\textsc{BR}& \textbf{0.16} & \textbf{0.25} & \textbf{0.26} & \textbf{0.37} & - & \textbf{0.75} & 0.35& (5-5) \\
\textsc{DR}& \textbf{0.05} & \textbf{0.11} & \textbf{0.09} & \textbf{0.11}& \textbf{0.25} & - & 0.12& (6-6) \\
\bottomrule
\end{tabular}
}
\caption{Win rates (WR) for each pair of systems for each of the three domains. The bold entries denote significance ($p < 0.05$) computed with Chi-square test. The ranking ranges are computed using bootstrap sampling. 
}
\label{tab:win_rates}
\end{table}
As expected, DR performs worst in all three domains, which is due to its repetitive nature, which is exposed over the course of a dialogue. In the Dailydialog and the Empathetic Dialogues domains, the GPT2 and the BR models perform equally, i.e., they end up in the same cluster. In both domains, systems using pre-trained language models outperform the S2 model, which is learned from scratch, which aligns with the expectation of related findings.
The BL model outperforms all other models in both the PersonaChat and Empathetic Dialogues domains, which is in line with the results presented by the authors of the Blender model \citep{roller2020recipes}. 
Furthermore, the LC model is ranked very highly. This corresponds to the findings of the ConvAI2 challenge \citep{dinan2020convai2}. However, in \emph{Spot The Bot}, the KV is ranked much higher than the HF model, which is not in line with the ConvAI2 evaluation.

\subsection{Survival Analysis}\label{sec:survival_results}

\begin{table}[h!]
    \centering
    \small
\begin{tabular}{l|ccc}
    
    \multicolumn{4}{c}{Dailydialog} \\
    \toprule
    & Fluency & Specificity & Sensibleness \\ \hline
    GPT & \bf 0.69 & 0.55     & \bf 0.77\\
    BR  & 0.77     & \bf 0.78 & \bf 0.62\\
    S2  & 0.31     & 0.52     & \bf 0.41\\
    DR  & \bf 0.23 & 0.15     & \bf 0.20\\
    \bottomrule
    

    \multicolumn{4}{c}{Empathetic Dialogues} \\
    \toprule
    & Fluency & Specificity & Sensibleness \\ \hline 
    BL  & 0.84     & 0.79 & \bf 0.84\\
    GPT & \bf 0.51 & 0.42 & \bf 0.49\\
    BR  & \bf 0.60 & 0.65 & \bf 0.56\\
    S2  & \bf 0.33 & 0.47 & \bf 0.39\\
    DR  & \bf 0.21 & 0.17 & \bf 0.21\\
    \bottomrule
    

    \multicolumn{4}{c}{PersonaChat} \\
    \toprule
    & Fluency & Specificity & Sensibleness \\ \hline
    BL  & \bf 0.73 & 0.74     & \bf 0.73\\
    LC  & 0.56     & 0.54     & \bf 0.62\\
    KV  & \bf 0.61 & 0.63     & \bf 0.58\\
    HF  & \bf 0.46 & \bf 0.46 & \bf 0.47\\    
    BR  & 0.48     & 0.44     & \bf 0.43\\
    DR  & \bf 0.16 & 0.19     & \bf 0.16\\
    \bottomrule

\end{tabular}
    \caption{Per feature win-rate of the different systems over all domains. Bold numbers indicate that the feature has a significant
    influence on system survival according to a Cox model.}
    \label{tab:survival_features}
\end{table}

\begin{figure*}[h!]
  \begin{subfigure}[b]{0.3\textwidth}
    \includegraphics[width=\textwidth]{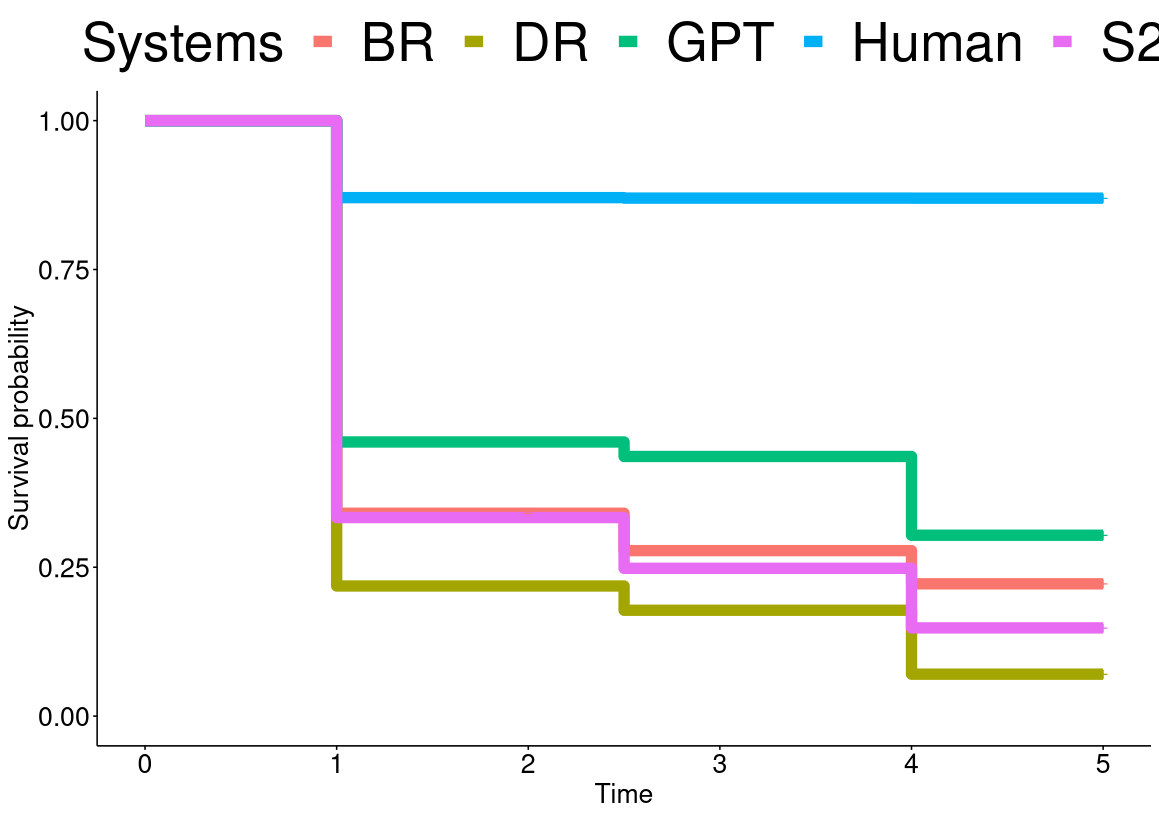}
    \caption{Dailydialog}
  \end{subfigure}
  \begin{subfigure}[b]{0.3\textwidth}
    \includegraphics[width=\textwidth]{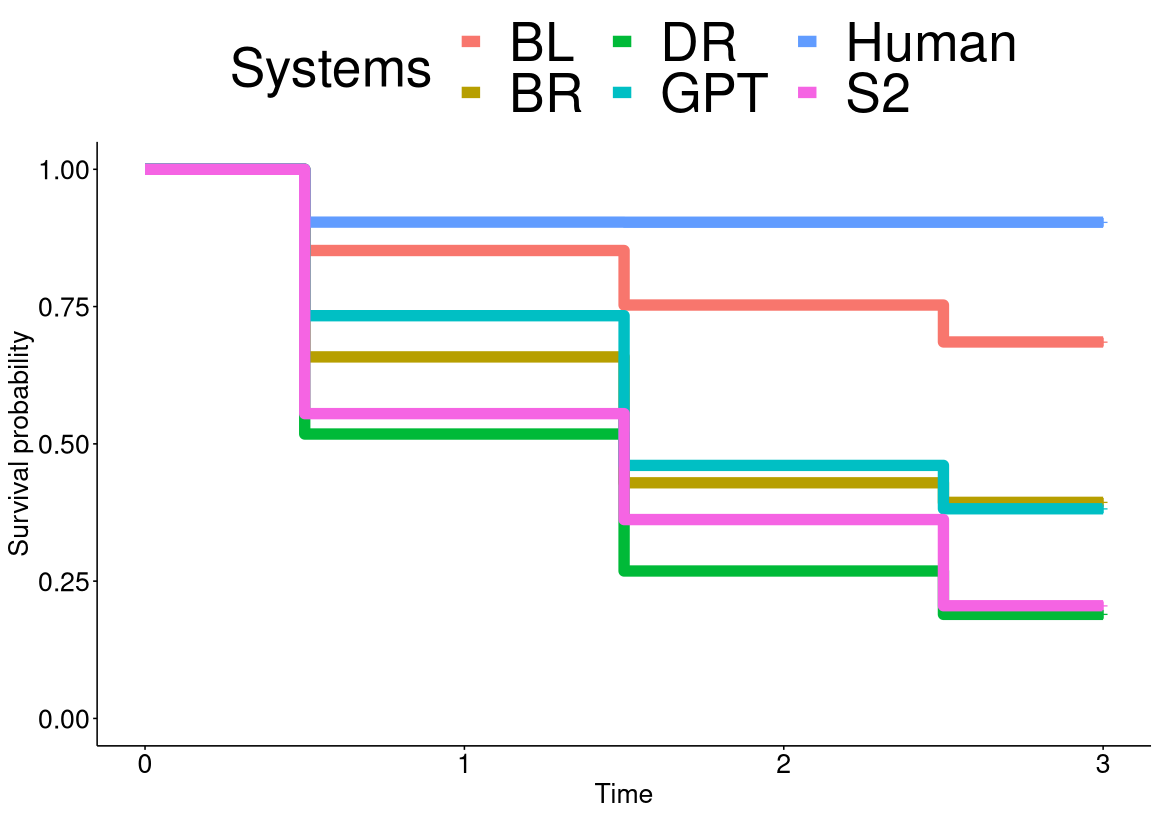}
    \caption{Empathetic Dialogues}
  \end{subfigure}
  \begin{subfigure}[b]{0.3\textwidth}
    \includegraphics[width=\textwidth]{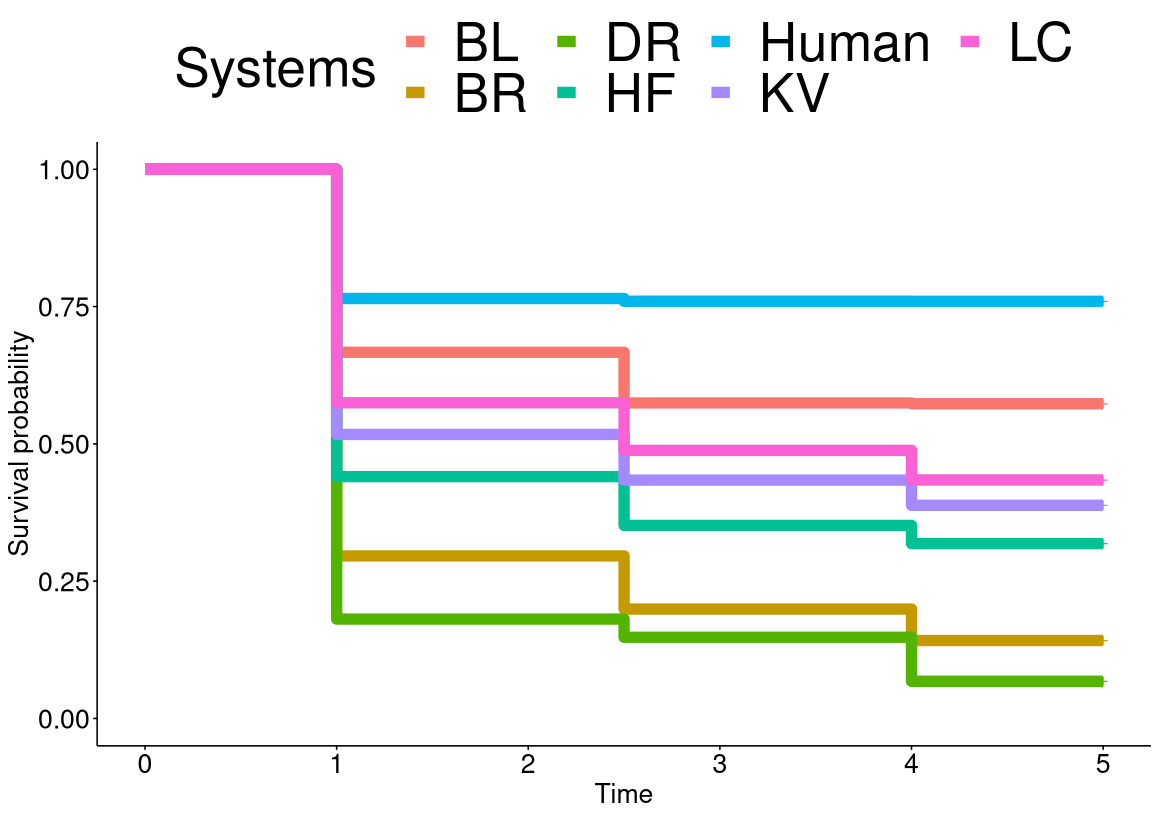}
    \caption{PersonaChat}
  \end{subfigure}
  \caption{Survival function per system estimated for each domain.}
  \label{fig:survival_all_domains}
\end{figure*}

Figure \ref{fig:survival_all_domains} shows the survival functions for the three domains. The survival rates produce the same rankings as those from pairwise win rates reported in Table \ref{tab:win_rates}, except for the Empathetic Dialogues domain, where GPT and BR switch places. Importantly, the distinction between these two is not significant in any of the rankings. Further non-significant differences within the Survival Analysis are S2 and DR in the Empathetic Dialogues domain, BR and S2 in the Dailydialog domain, and LC and KV in the PersonaChat domain. All other pairwise comparisons of survival curves are significant with $p  < 0.05$ after correction for multiple comparisons.\\

\noindent{\bf Feature Influence.} For each of the three features -- fluency, specificity, and sensibleness -- annotators have to specify whether one entity performed better, the same, or worse than the other. We encode this information as $1$, $0$, and $-1$ respectively and fit a Cox proportional hazards model \citep{cox_ph_model} for every system independently with the features as covariates.

The numerical entries in Table \ref{tab:survival_features} refer to the per-feature win-rate of each bot, which is computed analogously to Equation \ref{eq:win_function} using the feature annotations directly.
Bold entries in Table \ref{tab:survival_features} show which features have a significant influence on the system being spotted. All significant effects go in the intuitive direction, meaning that a higher feature value leads to longer survival.  For example, for the DR model, the fluency feature is significant across all three domains, and together with its low fluency win rate, we can deduce that it is often spotted due to its low fluency. Sensibleness seems to be an important feature across the board, meaning that in general, bots can be spotted due to inappropriate, nonsensical answers or hide if they respond in a suitable manner. Interestingly, specificity seems to be mostly unimportant, which could be due to either the bots not being noticeably unspecific, or it being an irrelevant feature for the chosen domains.

\section{Discussion}

\subsection{On Inter-Annotator Agreement}
The robustness of the evaluation of chatbots is often hampered by inter-annotator agreement (IAA) \cite{gandhe2016semi}. Measuring and reporting IAA is not yet a standard practice in evaluating chatbots \cite{amidei2019agreement}, and producing annotations with high IAA on open-domain conversations is prone to be impeded by subjective interpretation of feature definitions and idiosyncratic annotator behavior \cite{bishop2015use}. \\
In our setting, annotator disagreement on a bot's human-like behavior can be interpreted as a \emph{feature} of a bot's performance: A bot that manages to fool one of two annotators into believing it is human can be said to have performed better than a bot that does not manage to fool any annotator. \\
To analyze the annotator agreement in this light, we calculate per bot and label the percentage of cases where both annotators annotate the label if one of them does. Given three labels (\emph{human, bot, unsure}), the chance for random agreement is 0.33. 
The results averaged over all investigated domains and segment lengths per bot, are shown in Table \ref{tab:agreement_per_label_avg}.\footnote{We also analyzed agreement per segment length and domain but found no significant difference to averaging agreement over domains and segment lengths.} \\
\begin{table}[ht]
    \centering
    \small
    \begin{tabular}{l|ccc}
    
    \toprule
    label & bot $\downarrow$ & human $\uparrow$ & unsure \\ \hline

    \it human  & \it 0.33 & \it 0.84 & \it 0.15 \\
    BL  & 0.38 & 0.65 & 0.14 \\
    LC  & 0.60 & 0.52 & 0.10 \\
    GPT & 0.65 & 0.48 & 0.15 \\
    HF  & 0.70 & 0.41 & 0.10 \\
    KV  & 0.64 & 0.49 & 0.08 \\
    BR  & 0.74 & 0.39 & 0.15 \\
    DR  & 0.85 & 0.29 & 0.17 \\

    \bottomrule

    \end{tabular}
    \caption{Annotator agreement on labels.}
    \label{tab:agreement_per_label_avg}
\end{table}

The results confirm that the bots that rank high based on win rates and in the survival analysis (BL, GPT, LC) obtain the highest agreement on the \emph{human} label and lowest agreement on the \emph{bot} label. 
Conversely, the DR system obtains the highest agreement when being identified as a bot, and lowest when it is perceived as a human. \\
This analysis suggests that our experiments' results do not stem from a random agreement between the annotators, i.e., \ the annotations of the best and worst-performing systems show agreement distinctly higher than chance regarding the respective labels. 

\subsection{On Reliability}
One key requirement for an evaluation procedure is that repeated executions of the procedure result in the same outcome. We measure how many pairwise conversations between two bots are needed to guarantee a stable ranking. That is, what is the lower bound to $|S_{ij}|$ so that the ranking is stable. For each $|S_{ij}| \in \{3 ... 45\}$, we randomly sample $|S_{ij}|$ conversation for each pair and compute the ranking. We repeat this subsampling procedure 1000 times and measure the minimum $|S_{ij}|$ that guarantees the same ranking in at least $95\%$ of cases.
\begin{figure}[ht!]
  \begin{subfigure}[b]{0.45\textwidth}
    \includegraphics[width=\textwidth]{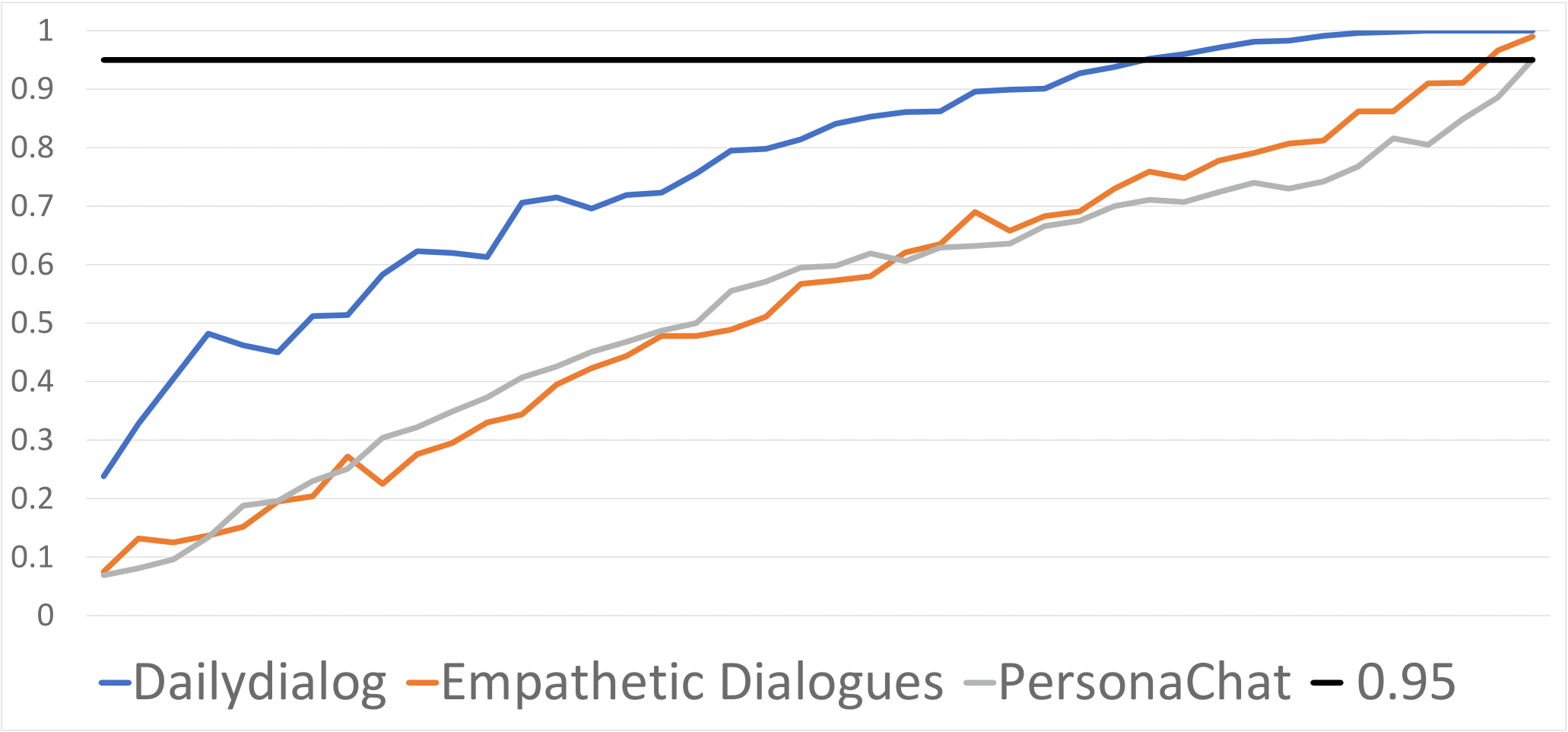}
    \caption{Stability Experiment.}
    \label{fig:ranking-significance}
  \end{subfigure}
  \begin{subfigure}[b]{0.45\textwidth}
    \includegraphics[width=\textwidth]{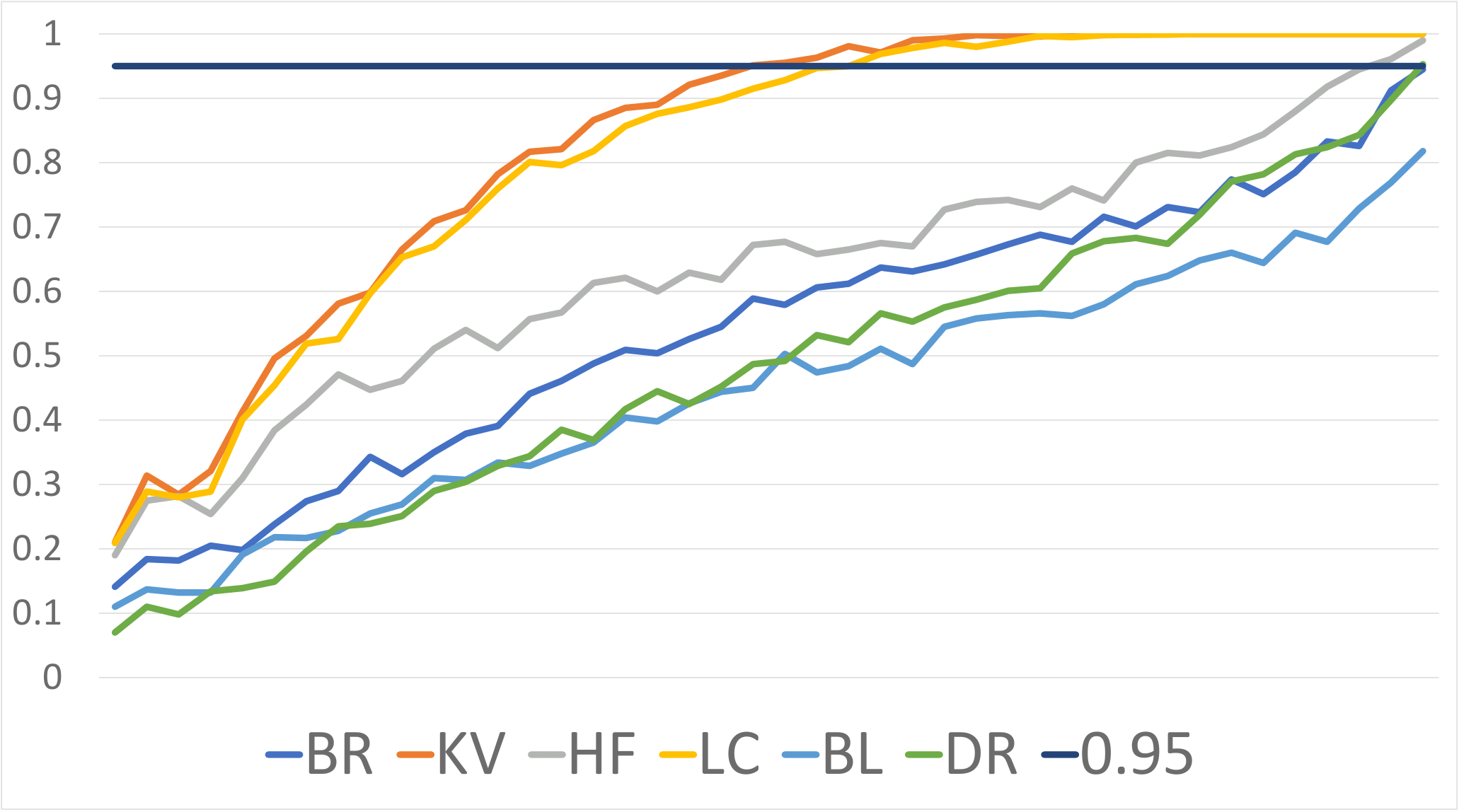}
    \caption{Leave-one-out Experiment.}
    \label{fig:ranking-significance-loo}
  \end{subfigure}
  \caption{Ranking stability experiments.The x-axis denotes the number of pairwise conversations between two bots. The y-axis denotes the rate at which the same ranking is achieved across 1000 repetitions. The horizontal line denotes the 95\% mark. In the lower Figure, we show the experiments for the PersonaChat domain, when leaving one system out.}
\end{figure}
Figure \ref{fig:ranking-significance} shows for each $|S_{ij}| \in \{3 ... 45\}$ the proportion of times in which the most frequent ranking occurred. For the Dailydialog domain, $|S_{ij}| = 33$ pairwise conversations are enough to guarantee a stable ranking. In the other two domains, this value is reached with over 40 pairwise dialogues. \\
A more in-depth analysis reveals that ranking stability depends on the significance of pairwise comparisons. For instance, in the PersonaChat domain, the KV and LC systems are not significantly different, which leads to two different rankings depending on the subsampling: in the first, KV and LC are in the same cluster, and in the second, LC and KV are in separate clusters, with LC being on top. Thus, removing either of them from the pool would yield a more stable ranking. To investigate this further, we applied a leave-one-out stability analysis. More precisely, we applied the analysis on $B \setminus \{\textit{sys}_i\}$, where $\textit{sys}_i \in B$. Figure \ref{fig:ranking-significance-loo} shows the result of the leave-one-out stability analysis. When leaving one between LC or KV out, the stability is achieved with 25 pairwise dialogues. When removing one of the other systems, the stability is reached with at least 40 dialogues. Thus, the number of pairwise bot-bot chats needed for \emph{Spot the Bot} evaluation depends on the pool of bots to be evaluated and should be determined empirically.

\subsection{On Time Efficiency}
Evaluation methods, which are costly and take up a long time, slow down the development cycle of dialogue systems. \emph{Spot The Bot} brings down the cost and time effort compared to other methods. 
\begin{table}[h!]
\small
\vspace{-1mm}
\begin{center}
\resizebox{.45\textwidth}{!}{
\begin{tabular}{l||ccccc} 
\hline
\textsc{Domain}& \makecell{Annotation \\ Time (Sec)}&  \makecell{Time per \\ Conversation (Sec)}\\
\hline 
\textsc{Dailydialog} & 26  & 153  \\
\textsc{Empathetic Dialouges} & 18  & 136  \\
\textsc{PersonaChat} & 24  & 238  \\
\hline
\end{tabular}
}
\end{center}
\caption{Overview of time efficiency in Seconds. Spot The Bot annotation versus creating human-bot conversations.}
\label{table:annotator-cost}
\vspace{-2mm}
\end{table}
In Table \ref{table:annotator-cost} the mean time per annotation is displayed.  For the Dailydialog and PersonaChat domain, the average annotation time is at around 25 seconds. For the Empathetic Dialogues, it is at 18 seconds, which is due to the shorter dialogues. We compare this to the time to create conversations between humans and bots. We recruited three dialogue system experts from our lab to interact with the systems. Each expert created 5 conversations with each system. The average times do not take into account the time needed to instruct the experts. For the Dailydialog and Empathetic Dialogues domains, it takes over 2 Minutes per conversation. \\
For PersonaChat, the time increased to almost 4 minutes. Similarly to our experts, the average time for a human-bot conversation in the wild evaluation of the ConvAI2 challenge\footnote{http://convai.io/data/} also lies at 4 minutes\footnote{We consider only conversations that have at least 10 turns, which is comparable to the setting of our experts.}. 
Considering the 100 dialogues per system used in ConvAI, the evaluation time would be 2,000 minutes per system. In Spot the Bot, 40 annotations times 24 seconds mean 16 minutes per pair of systems. Assuming a comparison between 5 systems, an approach based on human-bot annotations such as ConvAI would require 20 thousand minutes, while Spot the Bot would do with 0,16 thousand minutes\footnote{The amount of time needed by ConvAI grows linearly with the number of systems, while Spot the Bot (and ACUTE-EVAL) would grow quadratically. A pool of five systems seems reasonable for a research team, but even for larger pools (up to 51 systems) \emph{Spot the Bot} is still more efficient.}. \\
Concerning other methods based on self-talk, ACUTE-EVAL did not report the time per annotation, but they reported the time required to achieve significant results in PersonaChat, which is close to 30 minutes. Our method requires only 16 minutes (with 40 annotations). Thus, \emph{Spot The Bot} increases the annotation speed while reducing the human raters' mental strain.

\section{Conclusion}
In this work, we introduced \emph{Spot The Bot}, a robust and time-efficient approach for evaluating conversational dialogue systems. It is based on conversations between bots rated by humans with respect to the bots' ability to mimic human behavior. We show that \emph{Spot The Bot} yields robust and significant results while reducing the evaluation time compared to other evaluation frameworks. A team of researchers who would like to benchmark their system against four competing chatbots could do that for the cost of fewer than 3 hours of crowdsourced annotations. Spot the Bot facilitates developers making real progress based on frequent manual evaluations data, avoiding the use of noisy automatic metrics or once-in-a-year costly manual evaluations. 
We make the framework as well as the data publicly available. 

\section*{Acknowledgments}
This work has been partially funded by the LIHLITH project supported by the EU ERA-Net CHIST-ERA; the Swiss National Science Foundation [20CH21\_174237]; the Agencia Estatal de Investigación (AEI, Spain) projects PCIN-2017-118 and PCIN-2017-085; Basque Government IT1343-19.Jon Ander Campos enjoys a doctoral grant from the Spanish MECD.

\bibliography{emnlp2020}
\bibliographystyle{acl_natbib}
\newpage
\clearpage
\appendix

\section{Annotation Tool}
\label{sec:appendix}

\begin{figure*}[!ht]
	\begin{center}
        \begin{tabular}{@{}c@{}}
		\includegraphics[width=0.45\textwidth]{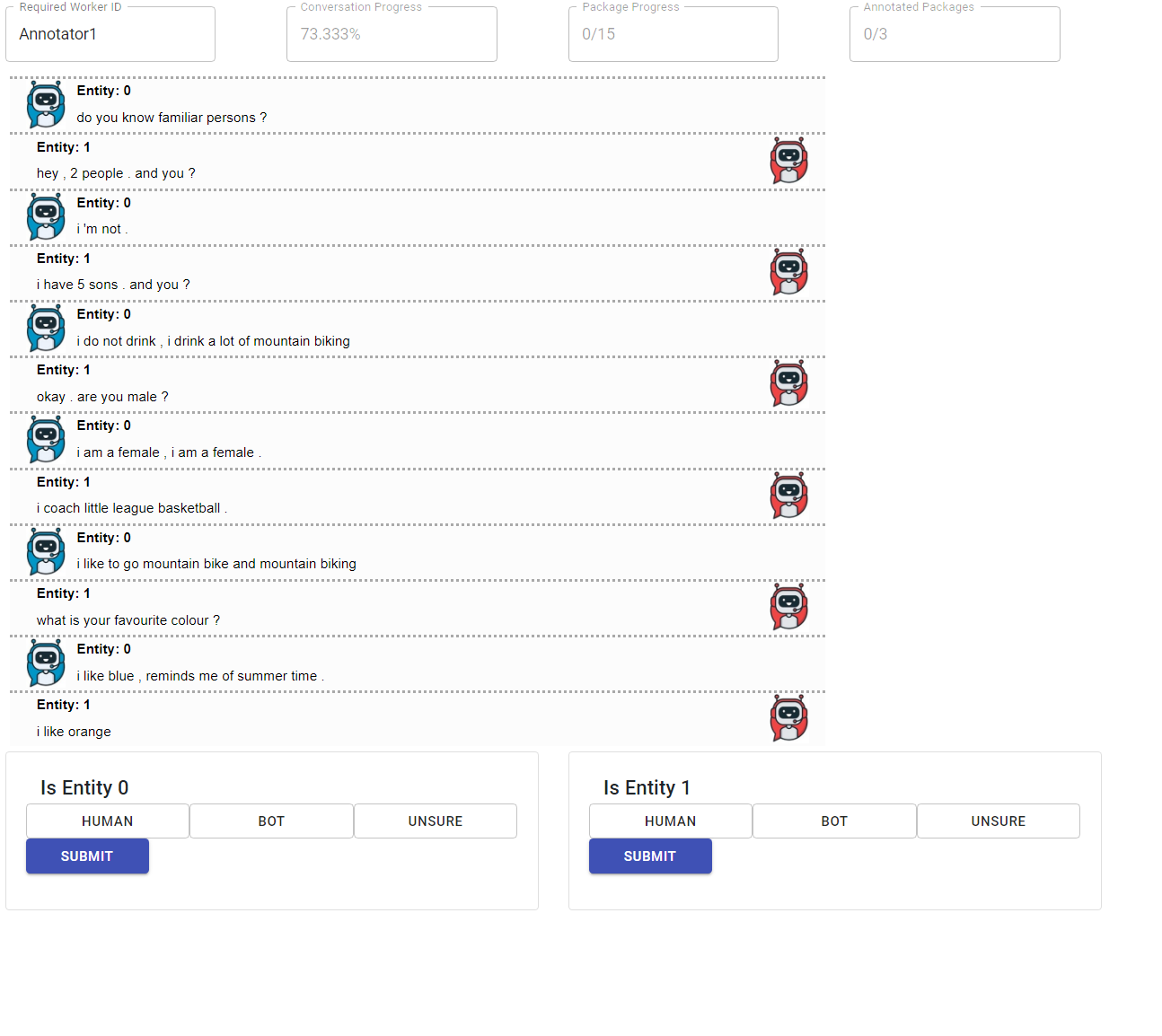} 
		\includegraphics[width=0.45\textwidth]{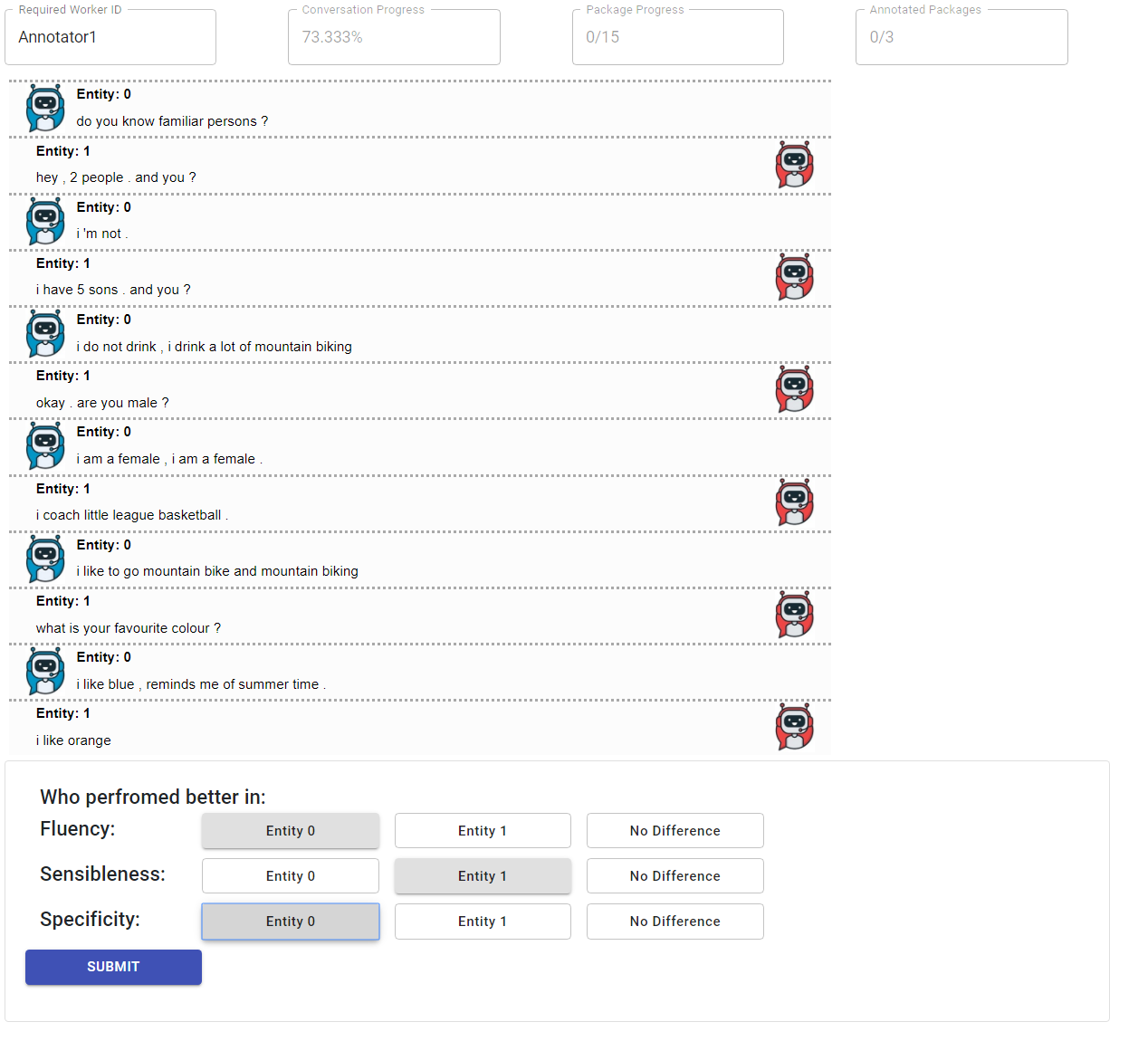}
       \end{tabular}
	\end{center}\vspace{-3mm}
	\captionof{figure}{The annotation tool. Left is the decision about the nature of each entity. Right is the decision with regard to the features. }
\label{fig:tool}
\end{figure*}

Figure \ref{fig:tool} shows the annotation tool. The annotator is presented with a segment of the conversation, with the first $i$ exchanges. In the first step, the annotator needs to decide for both entities separately if they are human or not. If it is not yet possible to decide, the annotator can choose to state that they are undecided. In the second step, the annotators are asked to state which of the two entities performs better with respect to three different features: fluency, sensibleness, and specificity with the following definitions:
\begin{itemize}
    \item Fluency: Which entities' language is more fluent and grammatically correct?
    \item Sensibleness: Which entities' responses are more sensible? If the answer seems confusing, illogical, contradictory, or factually wrong then it is NOT sensible.
    \item Specificity: Which entities' responses are more specific and explicit in the given context? An answer is specific if it can be given only in the current context.
\end{itemize}

\section{Gamification}
\label{sec:appendix_gamification}
\begin{figure*}[!ht]
	\begin{center}
        \begin{tabular}{@{}c@{}}
		\includegraphics[width=0.45\textwidth]{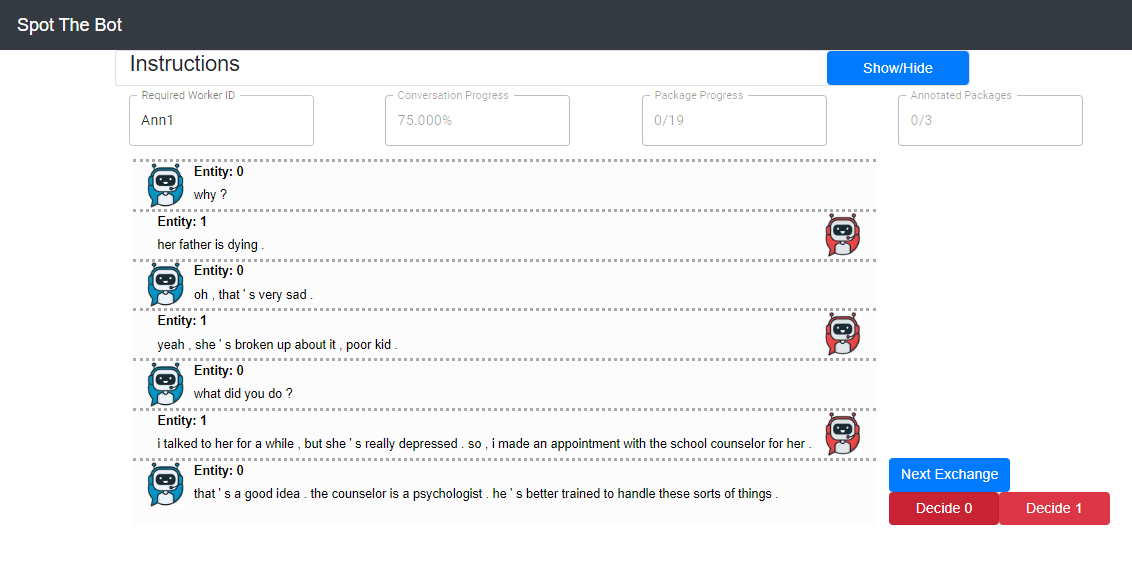} 
		\includegraphics[width=0.45\textwidth]{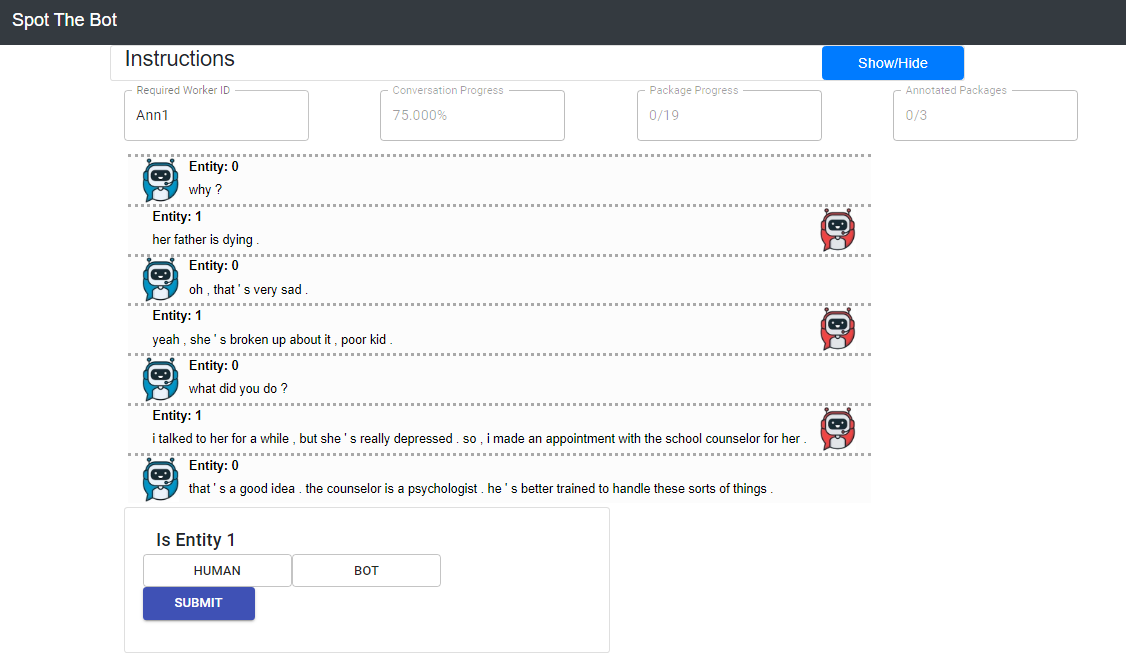}
       \end{tabular}
	\end{center}\vspace{-3mm}
	\captionof{figure}{Gamified version of the annotation tool. }
\label{fig:game-tool}
\end{figure*}

As an alternative to the segmentation approach, we experimented with a gamified version of the annotation tool (see Figure \ref{fig:game-tool}). In this version, the annotators were presented with the first turn of the conversation. At each point in time, they could choose whether to open the next turn or decide for an entity. If both decisions have been made, the annotators had to decide for the three aforementioned features, which entity performs better. 
The task was framed as a game, and the annotators received feedback in the form of a leaderboard. The score was a combination of the correctness (were the entities classified correctly) and a turn-penalty. That is, the more turns they opened, the lower the score. As an additional incentive, the winner was awarded a bonus payment. 
However, this approach resulted in unwanted behavior of the annotators. Some always decided after just one exchange, which leads to random annotations. Others opened the whole conversation first and then decided. To counteract these behaviors the tool needed a lot of fine-tuning, making the approach not reliable for practical use.

\section{Experimental Setup}
\label{app:experiments}
All the systems which we used were trained using the ParlAI system. We used the available models for the Lost in Conversation system, Blender, Huggingface system, and the KVMemNN. The other systems were trained using the ParlAI training functionality with the following hyperparameters. We trained all the models for 30 epochs.
For all the Bert-Rank experiments, we used the Bi-Encoder and optimized the last four layers due to GPU restrictions. 
The GPT2 models were trained with the standard-setting. Due to GPU restrictions, we used the small version of the GPT2 model. 
The sequence-to-sequence model was trained with two layers of GRUs \cite{cho2014gru}, each with 512 hidden units. We used the general attention mechanism \cite{luong-etal-2015-effective} and used the FastText word-embeddings\cite{bojanowski-etal-2017-enriching}. We used the ADAM optimizer \cite{kingma2014adam} with a learning rate of 0.001. 
For the small sequence-to-sequence model, we used a one layer GRU with 128 hidden units. We trained this model for only 3 epochs as we noted that after three epochs, it is able to generate the generic answers. 

\section{Feature Rankigns}
\begin{table}[h!]
\centering
\small
\scalebox{0.65}{
\begin{tabular}{l|cccccc|cc} 
\multicolumn{9}{c}{Dailydialog} \\
\toprule
\textsc{} & \textsc{GPT} &  \textsc{BR} &  \textsc{S2} & \textsc{DR} & & & \textsc{Win Rate}& \textsc{Range}\\
\hline 
\textsc{GPT} & -            & 0.54          & \textbf{0.85}    & \textbf{0.85} & & & 0.74 & (1,1) \\
\textsc{BR}  & 0.46         & -             & \textbf{0.79}    & \textbf{0.78} & & & 0.67 & (1,2)  \\
\textsc{S2} & \textbf{0.15} & \textbf{0.21} & -                 & \textbf{0.64} & & & 0.33 & (3,3) \\
\textsc{DR} & \textbf{0.15} & \textbf{0.22} & \textbf{0.36}     & -             & & & 0.24 & (4,4) \\
\bottomrule

\multicolumn{9}{c}{Empathetic Dialogues} \\
\toprule
\textsc{} & \textsc{BL} & \textsc{BR} &  \textsc{GPT} &  \textsc{S2} & \textsc{DR} & & \textsc{Win Rate} & \textsc{Range}\\
\hline 
\textsc{BL}   & -              & \textbf{0.72} & \textbf{0.86} & \textbf{0.85} & \textbf{0.94} && 0.84  & (1,1)  \\
\textsc{BR}   & \textbf{0.28}  & -             & 0.52          & \textbf{0.73} & \textbf{0.89} && 0.60  & (2,2)  \\
\textsc{GPT}  & \textbf{0.14}  & 0.48          & -             & \textbf{0.68} & \textbf{75}   && 0.51  & (2,3) \\
\textsc{S2}   & \textbf{0.15}  & \textbf{0.27} & \textbf{0.32} & -             & 0.59          && 0.33  & (4,4) \\
\textsc{DR}   & \textbf{0.06}  & \textbf{0.11} & \textbf{0.25} & 0.41          & -             && 0.19  & (5,5)\\
\bottomrule

\multicolumn{9}{c}{PersonaChat} \\
\toprule
\textsc{} & \textsc{BL}& \textsc{KV} &  \textsc{LC} &  \textsc{BR} & \textsc{HF}  & \textsc{DR} &\textsc{Win Rate} & \textsc{Range}\\
\hline 
\textsc{BL}& -             & \textbf{0.67} & \textbf{0.62} & \textbf{0.79} & \textbf{0.63} & \textbf{94}   &  0.73    & (1-1) \\
\textsc{KV}& \textbf{0.33} & -             & 0.54          & \textbf{0.66} & \textbf{0.70} & \textbf{0.83} & 0.61     & (2-3)\\
\textsc{LC}& \textbf{0.38} & 0.46          & -             & 0.52          & \textbf{0.60} & \textbf{0.83} & 0.56     & (2-4)\\
\textsc{BR}& \textbf{0.21} & \textbf{0.34} & 0.48          & -             & 0.61          & \textbf{0.78} & 0.48     & (3-5)\\
\textsc{HF}& \textbf{0.37} & \textbf{0.30} & \textbf{0.40} & 0.39          & -             & \textbf{0.82} & 0.45     & (3-5) \\
\textsc{DR}& \textbf{0.06} & \textbf{0.17} & \textbf{0.17} & \textbf{0.22} & \textbf{0.18} & -             & 0.16     & (6-6) \\
\bottomrule
\end{tabular}
}
\caption{Win rates for each pair of systems for each of the three domains. The bold entries denote significance ($p < 0.05$) computed with Chi-square test.}
\label{tab:fl-win-rates}
\end{table}

In Table \ref{tab:fl-win-rates}, the win rates and rankings for the fluency feature are shown. For the PersonaChat domain, the ranking differs significantly from the bot detection, as KV, LC, BR, and HF are all in the same cluster. 

\begin{table}[h!]
\centering
\small
\scalebox{0.65}{
\begin{tabular}{l|cccccc|cc} 
\multicolumn{9}{c}{Dailydialog} \\
\toprule
\textsc{} & \textsc{GPT} &  \textsc{BR} &  \textsc{S2} & \textsc{DR} & & & \textsc{Win Rate}& \textsc{Range}\\
\hline 
\textsc{GPT} & -            & 0.58          & \textbf{0.77}     & \textbf{0.86} & & & 0.74 & (1,1) \\
\textsc{BR}  & 0.42         & -             & \textbf{0.65}     & \textbf{0.87} & & & 0.64 & (2,2)  \\
\textsc{S2} & \textbf{0.23} & \textbf{0.35} & -                 & \textbf{0.76} & & & 0.44 & (3,3) \\
\textsc{DR} & \textbf{0.14} & \textbf{0.13} & \textbf{0.24}     & -             & & & 0.17 & (4,4) \\
\bottomrule

\multicolumn{9}{c}{Empathetic Dialogues} \\
\toprule
\textsc{} & \textsc{BL} & \textsc{BR} &  \textsc{S2} &  \textsc{GPT} & \textsc{DR} & & \textsc{Win Rate} & \textsc{Range}\\
\hline 
\textsc{BL}   & -              & \textbf{0.64} & \textbf{0.84} & \textbf{0.89} & \textbf{0.95} && 0.84  & (1,1)  \\
\textsc{BR}   & \textbf{0.36}  & -             & \textbf{0.63} & 0.56          & \textbf{0.94} && 0.62  & (2,2)  \\
\textsc{S2}  & \textbf{0.16}  & \textbf{0.37} & -             & 0.56          & \textbf{0.74} && 0.45   & (3,4) \\
\textsc{GPT} & \textbf{0.11}  &  0.44          & 0.44          & -             & \textbf{0.71} && 0.33  & (3,4) \\
\textsc{DR}   & \textbf{0.05}  & \textbf{0.06} & \textbf{0.26} & \textbf{0.29} & -             && 0.16  & (5,5)\\
\bottomrule

\multicolumn{9}{c}{PersonaChat} \\
\toprule
\textsc{} & \textsc{BL}& \textsc{KV} &  \textsc{LC} &  \textsc{HF} & \textsc{BR}  & \textsc{DR} &\textsc{Win Rate} & \textsc{Range}\\
\hline 
\textsc{BL}& -             & \textbf{0.71} & \textbf{0.62} & \textbf{0.72} & \textbf{0.84} & \textbf{0.94} & 0.76     & (1-1) \\
\textsc{KV}& \textbf{0.29} & -             & 0.56          & \textbf{0.73} & \textbf{0.70} & \textbf{0.89} & 0.63     & (2-3)\\
\textsc{LC}& \textbf{0.38} & 0.44          & -             & 0.57          & 0.55          & \textbf{0.85} & 0.56     & (2-3)\\
\textsc{HF}& \textbf{0.28} & \textbf{0.27} & 0.43          & -             & \textbf{0.63} & \textbf{0.81} & 0.48     & (4-4)\\
\textsc{BR}& \textbf{0.16} & \textbf{0.30} & 0.45          & \textbf{0.37} & -             & \textbf{0.76} & 0.41     & (4-5) \\
\textsc{DR}& \textbf{0.06} & \textbf{0.11} & \textbf{0.15} & \textbf{0.19} & \textbf{0.24} & -             & 0.15     & (6-6) \\
\bottomrule
\end{tabular}
}
\caption{Win rates for each pair of systems for each of the three domains. The bold entries denote significance ($p < 0.05$) computed with Chi-square test.}
\label{tab:ssa-win-rates}
\end{table}
In Table \ref{tab:ssa-win-rates} the win rates for the Sensibleness and
Specificity Average (SSA) are shown. A system wins if it is favored both in sensibleness and specificity. The rankings are similar to the bot detection rankings. For empathetic dialogues, the GPT model performs indistinguishably from the S2 model. In the PersonaChat domain, HF and BR are in the same cluster. 

\section{Domain Details}
\label{app:domain}

\begin{table}[h!]
\vspace{-1mm}
\begin{center}
\resizebox{.45\textwidth}{!}{
\begin{tabular}{l||cccl} 
\hline
\textsc{Domain Name} & \textsc{\#Dialogues} &  \textsc{Avg. Exchanges} &  \textsc{$|B|$} & \textsc{Segments}\\
\hline 
\textsc{Dailydialog}  &     13118   & 3.74 &   4  & 2,3,5 \\
\textsc{Empathetic Dialogues} &    25000   & 1.65 &   5   &  1,2,3   \\
\textsc{Personachat}    & 10907   & 7.85  &   6   &  2,3,5  \\
\hline
\end{tabular}
}
\end{center}
\caption{Overview of the domains}
\label{table:domain-overview}
\end{table}
We apply Spot The Bot on three different domains, which all are based on conversations between two humans. Thus, dialogue systems learn to imitate human conversational behavior. \\
\noindent{\bf Personachat.} PersonaChat \cite{zhang-etal-2018-personalizing} contains dialogues between two humans, each of the conversation participants is given a predefined persona. The persona is a set of characteristics of a person (name, occupation, hobbies, etc.), and the goal of the conversation is to mimic the process of getting to know each other. \\
\noindent{\bf Dailydialog.} Dailydialog \cite{li-etal-2017-dailydialog} is a dataset that contains dialogues that occur in daily life situations. The data is crawled from English learning websites. Thus, the dialogues are better curated and more formal. Furthermore, the data is annotated with features that represent the emotion in the dialogue. For our experiment, we did not make use of these features. \\
\noindent{\bf Empathetic Dialogues.} Empathetic Dialogues \cite{rashkin-etal-2019-towards} focuses on empathetic response generation. The dialogues occur between two persons that discuss a situation that happened to one of the participants. Thus, there are two types of participants: the speaker and the listener. The first describes the situation and their feelings about it, and the listener responds empathetically. 

\section{Segment Length Analysis}
\begin{table}[h!]
\centering
\small
\scalebox{0.95}{
\begin{tabular}{|l||cc|cc|cc|} 
\hline
\textsc{SYS/SEG} & \multicolumn{2}{c|}{2} &\multicolumn{2}{c|}{3}& \multicolumn{2}{c|}{5}\\
\hline 
 & WR & HP & WR & HP & WR & HP \\
\hline
\textsc{GPT}   & 0.75 & 0.30  & 0.75 & 0.34 & 0.81 & 0.22 \\
\textsc{BR}    & 0.60 & 0.22  & 0.64 & 0.21 & 0.70 & 0.15 \\
\textsc{S2}    & 0.46 & 0.20  & 0.39 & 0.17 & 0.34 & 0.11 \\
\textsc{DR}    & 0.16 & 0.11  & 0.20 & 0.11 & 0.13 & 0.04 \\
\hline
\textsc{Ties}  & \multicolumn{2}{c|}{72\%} &  \multicolumn{2}{c|}{75\% } &  \multicolumn{2}{c|}{81\%}\\
\hline
\end{tabular}
}
\caption{Segment Analysis for the Dailydialog domain. For each segment 2,3, and 5 the win-rate (WR) and the percentage of classification as humans (HP) are shown. In the last row the percentage of ties is shown.}
\label{tab:segment-analysis}
\end{table}
The intuition behind the segment length is that if the dialogue is too long, then most conversational dialogue systems will always be exposed as such. Contrary, if the dialogues are too short, there is too little information to discriminate between dialogue systems. Thus, having different lengths of conversations ensures that these extremes do not occur. The effect is shown in Table \ref{tab:segment-analysis}. For each dialogue system, the rate at which it is classified as a human is depicted for the three different segments. For each dialogue system, this rate goes down, which is in line with our intuition. Similarly, the rate of unsure classification is lower at later segments. In later segments, two phenomena occur. First, the number of ties increases, as most dialogue systems get exposed as such, the number of ties in the Dailydialog domain increases from 72\% to 81\%. Second, the difference between the win-rates increases. Better bots have a higher win-rate, and the lower-ranked bots get a lower win rate. However, the win-rates are less significant due to the high number of ties. For instance, the GPT model increases its win rate to 0.81, whereas the win rate for S2 decreases from 0.46 to 0.34.

\section{On Stability against weak Annotators}
One drawback of Likert-scale based evaluation methods is that many annotations need to be removed due to unreliable annotators \cite{lowe-etal-2017-towards}. \emph{Spot The Bot} shows that it is stable with respect to weak annotators. Since we can measure how often the annotators correctly classify an entity, we can rate the quality of an annotator. A random annotator would receive a correctness rate of 50\%. Table \ref{table:annotator-overview} shows an overview of the annotators for each domain. 

\begin{table}[h!]
\vspace{-1mm}
\begin{center}
\resizebox{.5\textwidth}{!}{
\begin{tabular}{l||cccc} 
\hline
\textsc{Domain} & \textsc{\#Ann} &  \textsc{Avg. Corr} & \textsc{Avg. Hum. Corr.} & $< 50\%$\\
\hline 
\textsc{DD} & 33   & 77\%  & 86\% & 9.1\%\\
\textsc{ED} & 32   & 63\%  & 92\% & 7.5\%\\
\textsc{PC} & 40  & 69\%  & 77\% & 22.8\%\\
\hline
\end{tabular}
}
\end{center}
\caption{Overview of the annotator performance. The number of annotations (\#Ann), the average correctness score (\textsc{Avg. Corr}), the average correctness score for the human-human conversations (\textsc{Avg. Hum. Corr.}), and the percentage of annotators that have a correctness score below 50\% ( $< 50\%$).}
\label{table:annotator-overview}
\end{table}
The average correctness score is significantly higher than random. For the Dailydialog and Empathetic Dialog domain, the rate of annotators, which achieved a rate below 50\%, was below 10\% of all annotators. For the PersonaChat domain, the rate is higher, which is due to the fact that stronger dialogue systems were in the pool of bots. The average correctness scores for predicting humans correctly is high for all domains. 
Hence, \emph{Spot The Bot} proves to be stable against annotators with low scores. 

When removing all annotators with scores below $75\%, $ the rankings remain stable. Only the significance scores decrease as a large number of dialogues gets removed. This lies in contrast to the gathering of conversations between humans and bots, which must be strictly supervised. For instance, the dialogues gathered in the wild evaluation of the ConvAI2 challenge were not usable. In fact, we applied Spot The Bot on these conversations, and the humans were rated as bots in 45\% of the cases.

\end{document}